\documentclass[a4paper]{svproc}

\usepackage[T1]{fontenc}

\usepackage{url}

\usepackage[noadjust]{cite}

\usepackage{amsmath,amssymb,mathtools}
\usepackage{booktabs}
\usepackage{array}
\usepackage{enumitem}

\newcommand{\R}{\mathbb{R}}
\newcommand{\E}{\mathbb{E}}
\newcommand{\Var}{\mathrm{Var}}
\newcommand{\tr}{\mathrm{tr}}

\newcommand{\HVP}{\mathrm{HVP}}

\begin{document}
\mainmatter

\title{Stochastic Estimation of the Layer-wise Hessian Trace
       for Monitoring Neural-network Training}
\titlerunning{Layer-wise Hessian-trace estimation}

\author{Maxim~A.~Bolshim \and Alexander~V.~Kugaevskikh}
\authorrunning{M.~A.~Bolshim, A.~V.~Kugaevskikh}
\tocauthor{Maxim A. Bolshim, Alexander V. Kugaevskikh}

\institute{ITMO University, St.~Petersburg, Russia\\
\email{maxim.bolshim@yandex.ru, a-kugaevskikh@yandex.ru}}

\maketitle

\begin{abstract}
The loss and the norm of its gradient separate the healthy and the
pathological regimes of neural-network training only weakly, whilst
the curvature of the empirical risk differs qualitatively between
them but is inaccessible explicitly at parameter counts
$P\sim 10^{6}\text{--}10^{8}$. We present a stochastic estimator of
the trace of the diagonal blocks of the Hessian matrix of the
empirical risk of a neural network. The
procedure combines the Hutchinson stochastic trace estimator with a
single Hessian--vector product over the whole parameter vector and
recovers unbiased estimates of every per-layer trace in one backward
pass through the computational graph. We show that correctness under
weight sharing requires the layer-wise Hessian to be assembled before
the second differentiation: unrolling shared weights into independent
coordinates introduces a systematic bias whose sign and magnitude are
governed by the cross-instance blocks of the unrolled Hessian. A
closed-form expression for the variance of the estimator at a fixed
Hessian is derived, together with a decomposition of the total variance
under the mini-batch sampling distribution. This decomposition yields a
critical probe count $K^{\star}$ that balances the two sources of
randomness and supports the practical recommendation
$K\in[5,10]$ in the on-line monitoring regime. The estimator is applied
to the detection of the label-memorisation regime of ResNet-18,
ResNet-34, and VGG-11 on CIFAR-10 and CIFAR-100, where a calibrated
cumulative-sum decision rule attains an empirical detection power of
$179/180$ at a false-alarm rate of $16/120$.

\end{abstract}

\section{Introduction}

The empirical loss and the norm of its gradient are functionals of the
first-order derivatives of the model, and several training regimes of
practical interest---memorisation of corrupted
labels~\cite{zhang2017understanding,arpit2017closer}, latent
instability, and the divergence of training from test error---are
weakly separable from the nominally healthy regime when viewed through
these first-order quantities alone. The curvature of the loss, encoded
in the Hessian matrix $H=\nabla^2_\theta\mathcal{L}(\theta)$, exhibits a
qualitatively different response in such
regimes~\cite{sagun2017empirical,yao2020pyhessian,papyan2020traces};
however, the explicit storage of $H$ is infeasible whenever the
parameter count satisfies $P\sim 10^{6}\text{--}10^{8}$.

The standard remedy is the stochastic estimation of scalar functionals
of $H$ via the Hutchinson
estimator~\cite{hutchinson1990stochastic,avron2011randomized} combined
with the Hessian--vector product of
Pearlmutter~\cite{pearlmutter1994fast}. A layer-wise resolution of the
curvature is natural: the Hessian is block-structured with respect to
the partition of the parameter vector into layers, and the diagonal
blocks $H_\ell$ describe the curvature along the parameters of a single
layer. Existing spectral
diagnostics~\cite{yao2020pyhessian,sagun2017empirical,ghorbani2019investigation,papyan2020traces}
act on the aggregated Hessian or on its global spectrum and do not
isolate the layer-wise estimation error; the principal second-order
approximations---the Becker--LeCun
diagonal~\cite{beckerlecun1988second}, K-FAC~\cite{martens2015optimizing}
and the Gauss--Newton
family~\cite{schraudolph2002fast,martens2010deep}---either disregard
within-layer correlations, suppress the negative curvature, or are
biased on strongly non-linear layers.

The present paper formalises a layer-wise stochastic estimator of
$\tr(H_\ell)$ that remains correct under weight sharing, derives a
closed-form expression for its variance and a decomposition of the
total variance in the mini-batch regime, and demonstrates the
applicability of the estimator on an on-line detector of the
label-memorisation regime with a calibrated false-alarm rate. A
companion preprint~\cite{bolshim_dag_preprint} develops the cross-layer
curvature formalism for general directed-acyclic-graph architectures.

\section{Problem statement and notation}

Let $f_\theta:\R^{d_0}\to\R^{m}$ denote a neural network parametrised
by $\theta\in\R^{P}$ partitioned into $L$ disjoint groups
$\theta=(\theta_1,\dots,\theta_L)$ with $\theta_\ell\in\R^{P_\ell}$ and
$\sum_\ell P_\ell=P$. The group $\theta_\ell$ collects all parameters
of the $\ell$-th layer, including those of weight-shared modules. The
empirical risk on a mini-batch $B\subset\mathcal{D}$ is
\begin{equation}
\label{eq:loss}
   \mathcal{L}_B(\theta)
   = \frac{1}{|B|}\!\sum_{(x,y)\in B}\!\ell\!\bigl(f_\theta(x),y\bigr),
   \qquad
   \mathcal{L}(\theta) = \E_{B}\,\mathcal{L}_B(\theta).
\end{equation}
The full Hessian is $H(\theta)=\nabla^2_\theta\mathcal{L}(\theta)
\in\R^{P\times P}$ and $H_B(\theta)$ denotes its mini-batch
realisation.

\paragraph{Layer-wise Hessian and target quantity.}
The diagonal block and its trace are
\begin{equation}
\label{eq:layer-hessian}
   H_\ell(\theta) := \nabla^2_{\theta_\ell}\mathcal{L}(\theta)
   \in \R^{P_\ell\times P_\ell},
   \quad
   T_\ell(\theta) := \tr\bigl(H_\ell(\theta)\bigr)
                  = \sum_{i=1}^{P_\ell}\lambda_i(H_\ell).
\end{equation}
The matrix $H_\ell$ is symmetric but is not positive semi-definite in
general. The mini-batch realisation
$T_{\ell,B}:=\tr(H_{\ell,B})$ is unbiased in the sense
$\E_B\,T_{\ell,B}=T_\ell$.

\paragraph{Hessian--vector product.}
For $v\in\R^{P_\ell}$ the product $H_\ell v$ is computed without
explicitly forming $H_\ell$ by the method of
Pearlmutter~\cite{pearlmutter1994fast}, namely by differentiating the
scalar $\langle\nabla_{\theta_\ell}\mathcal{L},v\rangle$ a second time.
The cost is of the order of two backward passes and is independent of
$P_\ell$; it is denoted by $\mathcal{C}_\HVP$. When $v\in\R^{P}$, a
single call to automatic differentiation delivers $H_\ell v_\ell$
simultaneously for every $\ell$ at the same cost $\mathcal{C}_\HVP$;
this property is exploited in Section~\ref{sec:algorithm}. The cost of
a single gradient pass is denoted by $\mathcal{C}_\nabla$.

\paragraph{Random probes.}
In both regimes considered below ($z\in\R^{P_\ell}$ in
Section~\ref{sec:base-estimator} and $z\in\R^{P}$ in
Section~\ref{sec:algorithm-statement}) the probes satisfy
\begin{equation}
\label{eq:probe-conditions}
   \E[z] = 0, \qquad \E[z z^{\!\top}] = I.
\end{equation}
The canonical choice is Rademacher, namely $z_i\in\{-1,+1\}$
independently and equiprobably across coordinates; it attains the
minimum variance of the Hutchinson estimator of the trace of a
symmetric matrix within the class of distributions with independent
coordinates that satisfy~\eqref{eq:probe-conditions}%
~\cite{avron2011randomized}.

\section{Algorithm for the layer-wise trace}
\label{sec:algorithm}

\subsection{Basic estimator}
\label{sec:base-estimator}

For a fixed matrix $H_\ell\in\R^{P_\ell\times P_\ell}$, the Hutchinson
estimator of its trace based on $K$ independent probes
$z_1,\dots,z_K\in\R^{P_\ell}$ satisfying~\eqref{eq:probe-conditions} is
defined by
\begin{equation}
\label{eq:hutchinson}
   \hat T_\ell \;:=\; \frac{1}{K}\sum_{k=1}^{K} z_k^{\!\top} H_\ell\,z_k.
\end{equation}
The estimator is unbiased:
$\E[z^{\!\top}\!H_\ell z]
   = \tr\bigl(\E[zz^{\!\top}]H_\ell\bigr) = T_\ell$,
and hence $\E[\hat T_\ell]=T_\ell$ for every probe distribution
satisfying~\eqref{eq:probe-conditions}.

\subsection{Correctness under weight sharing}
\label{sec:weight-sharing}

For convolutional, recurrent, and normalising layers the parameter
group $\theta_\ell$ enters $M$ separate invocations of the layer.
Denote the virtual copies of the shared weights by
$W^{(1)},\dots,W^{(M)}\in\R^{P_\ell}$ with $W^{(k)}=\theta_\ell$, and
let $\tilde{\mathcal{L}}(W^{(1)},\dots,W^{(M)})$ be the loss expressed
through these copies. The chain rule yields
\begin{equation*}
   \nabla_{\theta_\ell}\mathcal{L}
   \;=\; \sum_{k=1}^{M}
         \frac{\partial\tilde{\mathcal{L}}}{\partial W^{(k)}},
\end{equation*}
and a further differentiation with respect to $\theta_\ell$ gives
\begin{equation}
\label{eq:shared-hessian}
   H_\ell(\theta_\ell)
   \;=\; \sum_{k,k'=1}^{M}
         \frac{\partial^{2}\tilde{\mathcal{L}}}
              {\partial W^{(k)}\,\partial W^{(k')}}.
\end{equation}
The product $H_\ell v$ for $v\in\R^{P_\ell}$ is therefore delivered by
a single application of Pearlmutter's
method~\cite{pearlmutter1994fast} to the scalar
$\langle\nabla_{\theta_\ell}\mathcal{L},v\rangle$, evaluated by
automatic differentiation with the computational graph retained.
Unrolling the shared weights into independent coordinates prior to the
second differentiation discards the cross-instance terms $k\neq k'$
in~\eqref{eq:shared-hessian} and produces an estimator with a
systematic bias whose sign and magnitude are governed by the spectrum
of the off-diagonal blocks
$\partial^{2}\tilde{\mathcal{L}}/
 \partial W^{(k)}\partial W^{(k')}$, and are otherwise unconstrained.

\subsection{The algorithm}
\label{sec:algorithm-statement}

Let $z\in\R^{P}$ satisfy~\eqref{eq:probe-conditions} in the ambient
dimension $P$, so that $\E[z_\ell z_m^{\!\top}]=\delta_{\ell m}I$. For
$m\neq\ell$ the independence of $z_\ell$ and $z_m$ implies
$\E[z_\ell^{\!\top}H_{\ell m}\,z_m]=0$, and consequently the
$\ell$-block $w_\ell$ of the vector $w=Hz$ obtained by a single call
to automatic differentiation satisfies
$\E[z_\ell^{\!\top} w_\ell]=T_\ell$: an unbiased estimator of the
trace of every layer is obtained without separate backward passes per
layer. The cost of $K$ probes across the whole network is
$K\,\mathcal{C}_\HVP$ rather than $LK\,\mathcal{C}_\HVP$.

The combination of the basic estimator with the single-pass scheme
yields the following procedure. The input is a model $f_\theta$, a
mini-batch $B$, and a probe count $K$; the output is the set of
estimates $\hat T_1,\dots,\hat T_L$ of the traces of the diagonal
blocks.

\begin{enumerate}[topsep=2pt,itemsep=1pt,parsep=0pt,leftmargin=*]
\item Compute the gradient $g\gets\nabla_\theta\mathcal{L}_B(\theta)$
      by a single backward pass with the computational graph retained;
      initialise $s_\ell\gets 0$ for every $\ell$.
\item For $k=1,\dots,K$, draw $z^{(k)}\in\R^{P}$ with independent
      Rademacher coordinates, compute
      $w^{(k)}\gets\nabla_\theta\langle g,z^{(k)}\rangle$ (the
      Hessian--vector product $H z^{(k)}$ obtained by a single call
      to automatic differentiation), and for every $\ell$ accumulate
      $s_\ell\mathrel{+}=\langle z_\ell^{(k)},w_\ell^{(k)}\rangle$.
\item Set $\hat T_\ell\gets s_\ell/K$ for $\ell=1,\dots,L$.
\end{enumerate}

\noindent
The total cost is $\mathcal{C}_\nabla+K\mathcal{C}_\HVP$ and the
memory footprint is $O(P)$ for the storage of $g$, $z^{(k)}$, and
$w^{(k)}$. A justification of the recommended value
$K\in[5,10]$ is given in Section~\ref{sec:variance}.

\section{Variance analysis}
\label{sec:variance}

\subsection{Variance at a fixed Hessian}
\label{sec:variance-fixed}

For a symmetric matrix $H_\ell\in\R^{P_\ell\times P_\ell}$ and $K$
independent Rademacher probes $z_1,\dots,z_K$, a direct evaluation of
$\E[(z^{\!\top}\!H_\ell z)^2]$ using the moment identity
\begin{equation*}
   \E[z_iz_jz_kz_l] \;=\;
   \delta_{ij}\delta_{kl}+\delta_{ik}\delta_{jl}
   +\delta_{il}\delta_{jk}-2\delta_{ijkl}
\end{equation*}
yields
\begin{equation}
\label{eq:variance-hutchinson}
   \Var\!\bigl(\hat T_\ell \,\big|\, H_\ell\bigr)
   \;=\; \frac{2}{K}\!\left(\|H_\ell\|_F^{2}
   - \sum_{i=1}^{P_\ell}(H_\ell)_{ii}^{2}\right).
\end{equation}
The factor $1/K$ arises from the independence of the probes.

\subsection{Relative error and anisotropy}
\label{sec:anisotropy}

Define the anisotropy of the layer-wise Hessian by
\begin{equation}
\label{eq:anisotropy}
   \kappa_\ell \;:=\; \frac{\|H_\ell\|_F}{|T_\ell|}.
\end{equation}
From~\eqref{eq:variance-hutchinson} and
$\sum_i(H_\ell)_{ii}^{2}\ge 0$,
$\Var(\hat T_\ell\,|\,H_\ell)\le(2/K)\|H_\ell\|_F^{2}$;
for $T_\ell\neq 0$,
\begin{equation}
\label{eq:rel-error-bound}
   \frac{\sqrt{\Var(\hat T_\ell\,|\,H_\ell)}}{|T_\ell|}
   \;\le\; \sqrt{\frac{2}{K}}\;\kappa_\ell.
\end{equation}
In terms of the eigenvalues $\lambda_i$ of $H_\ell$ one has
\begin{equation*}
   \kappa_\ell
   \;=\; \Bigl(\textstyle\sum_i\lambda_i^{2}\Bigr)^{1/2}
         \Big/\Bigl|\textstyle\sum_i\lambda_i\Bigr|;
\end{equation*}
for a positive semi-definite matrix of rank $r$ the
Cauchy--Schwarz inequality gives $1/\sqrt{r}\le\kappa_\ell\le 1$,
whilst for an indefinite matrix $\kappa_\ell$ can be made arbitrarily
large by a mutual cancellation of eigenvalues of opposite sign in the
trace. Layers with $\kappa_\ell\gg 1$ require either a larger probe
count $K$ or a different target functional.

\subsection{Variance decomposition in the mini-batch regime}
\label{sec:variance-batch}

In practice $\hat T_\ell$ is constructed from the mini-batch Hessian
$H_{\ell,B}$. Under the independence of $B$ and of the probes the law
of total variance gives
$\Var(\hat T_{\ell,B}) = V_{\mathrm H}(K) + V_{\mathrm B}$,
where $V_{\mathrm H}(K)=\E_B[\Var_z(\hat T_{\ell,B}\,|\,B)]\propto 1/K$
by~\eqref{eq:variance-hutchinson} and
$V_{\mathrm B}=\Var_B(T_{\ell,B})$ is independent of $K$. The
critical probe count
\begin{equation}
\label{eq:K-star}
   K_\ell^{\ast}
   \;:=\; \frac{V_{\mathrm H}(1)}{V_{\mathrm B}}
   \;=\; \frac{2\,\E_B\!\bigl[\|H_{\ell,B}\|_F^{2}
         - \sum_i(H_{\ell,B})_{ii}^{2}\bigr]}
              {\Var_B\!\bigl(T_{\ell,B}\bigr)}
\end{equation}
balances the contributions of the two sources of randomness: for
$K\ge K_\ell^{\ast}$ the marginal benefit of increasing $K$ for the
estimation of the target $\E_B T_{\ell,B}$ becomes negligible. A
numerical evaluation of $K_\ell^{\ast}$ for ResNet-18 and VGG-11 on
CIFAR-10 with $|B|=128$~\cite{bolshim_dag_preprint} delivers values of
order $5\text{--}10$, which justifies the choice $K=10$ adopted in the
on-line monitoring regime.

\section{Illustration: monitoring the memorisation regime}
\label{sec:illustration}

The applicability of the proposed estimator is illustrated on the
classical task in which the loss and the gradient norm separate the
healthy and the pathological regimes only weakly, whilst the curvature
differs qualitatively.

\subsection{Experimental setup}
\label{sec:setup}

The datasets are CIFAR-10 and
CIFAR-100~\cite{krizhevsky2009cifar} with symmetric label noise at
intensity $\eta\in\{0,\,0.25,\,0.40,\,0.60\}$, injected prior to
training; the test set is kept clean. The architectures are
ResNet-18, ResNet-34~\cite{he2016deep} and
VGG-11~\cite{simonyan2015vgg}; optimisation is by SGD with momentum
$0.9$, base learning rate $0.1$, mini-batch $|B|=128$, a cosine
schedule, and ten random seeds per configuration. An $L_2$ regulariser
with coefficient $\lambda=5\times 10^{-4}$ is used; this is equivalent
to the addition of a constant term $2\lambda I$ to the diagonal of the
layer-wise Hessian and is taken into account in the interpretation of
$\hat T_\ell$. No gradient clipping is applied. The estimator
$\hat T_\ell$ is evaluated with the normalisation layers in
\texttt{eval} mode, which corresponds to $\nabla^2_\theta\mathcal{L}$
at fixed BatchNorm running statistics. The trajectory $\hat T_\ell(t)$
is computed by the procedure of
Section~\ref{sec:algorithm-statement} with $K=10$ every $100$
optimisation steps.

The Phase~I calibration ensemble comprises $S=20$ runs on clean labels
($\eta=0$) with independent initialisations for every pair
(architecture, dataset). From this ensemble the baseline statistics
$\mu_0(t)$ and $\sigma_0(t)$ of the trajectory $\hat T_\ell(t)$ are
obtained; they enter the Phase~II decision rule.

\subsection{Metrics and decision rule}
\label{sec:metrics}

For every layer $\ell$ and noise level $\eta$ the trajectories
$\{\hat T_\ell^{(s)}(t)\}_{s=1}^{10}$ are recorded. The effect size
between the clean and the noisy regimes is given by Cohen's~$d$,
\begin{equation}
\label{eq:cohens-d}
   d_\ell(\eta) \;=\; \frac{\mu_0(\ell) - \mu_\eta(\ell)}
   {\sqrt{\bigl(s_0^{2}(\ell) + s_\eta^{2}(\ell)\bigr)/2}},
\end{equation}
where $\mu_\eta$ and $s_\eta^{2}$ are the mean and the sample
variance of $\hat T_\ell$ on a fixed window of epochs $[10,20]$, and
$95\%$ confidence intervals are obtained by bootstrap.

The trajectory $\hat T_\ell(t)$ is non-stationary by construction
(the training step itself, the learning-rate schedule
$\eta_{\mathrm{lr}}(t)$, and the BatchNorm dynamics), and consequently
a decision rule with a fixed baseline window loses its consistency. A two-sided Page--Hinkley
cumulative-sum statistic~\cite{page1954cusum,hawkins1998cusum} is
employed instead. Standardisation against the ensemble,
$z_t = (\hat T_\ell(t) - \mu_0(t))/\sigma_0(t)$, is combined with the
recursion
\begin{equation}
\label{eq:cusum}
  S_t^{+} = \max\!\bigl(0,\, S_{t-1}^{+} + z_t - k\bigr),\qquad
  S_t^{-} = \max\!\bigl(0,\, S_{t-1}^{-} - z_t - k\bigr),
\end{equation}
with $S_0^{\pm}=0$ and $k=0.5$ (optimal for a shift of the order of
$1\sigma$). An alert is raised whenever $\max(S_t^{+},S_t^{-})>h$. The
threshold $h$ is calibrated so that the empirical mean run length to a
false alarm ($\mathrm{ARL}_0$) over the ensemble of clean runs equals
$1000$ snapshot steps---the standard construction in statistical
process control~\cite{montgomery2009spc}; the calibration is performed
by leave-one-out resampling on the Phase~I ensemble and bisection
in~$h$.

\subsection{Expected separability}
\label{sec:separability}

In the label-memorisation regime the layer-wise trace $\hat T_\ell$
at the classification head grows in comparison with the clean
regime~\cite{arpit2017closer,zhang2017understanding}; under the sign
convention of~\eqref{eq:cohens-d} this gives $d_\ell(\eta)<0$ and
triggers the rule~\eqref{eq:cusum}. At $\eta=0.60$ the model does not
consistently enter the memorisation regime, and the sign of the effect
is not guaranteed. The full numerical values of $d_\ell(\eta)$ are
reported in the companion preprint~\cite{bolshim_dag_preprint}.

\subsection{Empirical validation of the decision rule}
\label{sec:empirical}

The decision rule~\eqref{eq:cusum} is applied to all $300$ runs of
the testbed (six architecture--dataset configurations; $S=20$ clean
and $3\times 10$ noisy seeds per configuration). The threshold $h$ is
calibrated within each configuration by leave-one-out resampling at
$\mathrm{ARL}_0=1000$ snapshot steps with $k=0.5$. The outcome is
summarised in Table~\ref{tab:alerts}.

\begin{table}[!ht]
\centering
\footnotesize
\renewcommand{\arraystretch}{0.95}
\caption{Decision rule~\eqref{eq:cusum}: detection epoch
(mean~$\pm$~standard deviation, restricted to the seeds on which the
alert was raised), alert rate, and the calibrated threshold~$h$. The
row $\eta=0$ is the leave-one-out control and reports the empirical
false-alarm rate.}
\label{tab:alerts}
\begin{tabular}{lllccc}
\toprule
Arch & Dataset & $\eta$ & Det.\ epoch (mean$\pm$std) & Alert rate & $h$ \\
\midrule
resnet18 & cifar10 & $0.00$ & 16.5 $\pm$ 2.1 & 0.10 & 7.33 \\
resnet18 & cifar10 & $0.25$ & 22.0 $\pm$ 6.8 & 1.00 & 7.33 \\
resnet18 & cifar10 & $0.40$ & 2.7 $\pm$ 0.5 & 1.00 & 7.33 \\
resnet18 & cifar10 & $0.60$ & 2.1 $\pm$ 0.3 & 1.00 & 7.33 \\
resnet18 & cifar100 & $0.00$ & 25.0 $\pm$ 5.7 & 0.10 & 7.16 \\
resnet18 & cifar100 & $0.25$ & 1.9 $\pm$ 0.3 & 1.00 & 7.16 \\
resnet18 & cifar100 & $0.40$ & 1.3 $\pm$ 0.5 & 1.00 & 7.16 \\
resnet18 & cifar100 & $0.60$ & 1.1 $\pm$ 0.3 & 1.00 & 7.16 \\
resnet34 & cifar10 & $0.00$ & 3.5 $\pm$ 1.9 & 0.20 & 8.38 \\
resnet34 & cifar10 & $0.25$ & 21.0 $\pm$ 9.0 & 1.00 & 8.38 \\
resnet34 & cifar10 & $0.40$ & 12.9 $\pm$ 11.8 & 1.00 & 8.38 \\
resnet34 & cifar10 & $0.60$ & 3.1 $\pm$ 0.3 & 1.00 & 8.38 \\
resnet34 & cifar100 & $0.00$ & 18.5 $\pm$ 16.3 & 0.10 & 7.11 \\
resnet34 & cifar100 & $0.25$ & 1.9 $\pm$ 0.3 & 1.00 & 7.11 \\
resnet34 & cifar100 & $0.40$ & 1.8 $\pm$ 0.4 & 1.00 & 7.11 \\
resnet34 & cifar100 & $0.60$ & 1.2 $\pm$ 0.4 & 1.00 & 7.11 \\
vgg11 & cifar10 & $0.00$ & 13.0 $\pm$ 11.3 & 0.10 & 7.34 \\
vgg11 & cifar10 & $0.25$ & 11.1 $\pm$ 11.7 & 0.90 & 7.34 \\
vgg11 & cifar10 & $0.40$ & 3.2 $\pm$ 0.6 & 1.00 & 7.34 \\
vgg11 & cifar10 & $0.60$ & 3.0 $\pm$ 0.0 & 1.00 & 7.34 \\
vgg11 & cifar100 & $0.00$ & 3.5 $\pm$ 1.0 & 0.20 & 7.11 \\
vgg11 & cifar100 & $0.25$ & 2.1 $\pm$ 0.3 & 1.00 & 7.11 \\
vgg11 & cifar100 & $0.40$ & 2.0 $\pm$ 0.0 & 1.00 & 7.11 \\
vgg11 & cifar100 & $0.60$ & 2.0 $\pm$ 0.0 & 1.00 & 7.11 \\
\bottomrule
\end{tabular}

\end{table}

The power of the rule in the noisy regimes attains $1.0$ in $17$ of
the $18$ noisy cells (the single exception being vgg11/cifar10 at
$\eta=0.25$, with $0.90$); the pooled empirical power is
$179/180\approx 0.994$. The mean detection epoch decreases
monotonically with $\eta$ in every one of the six configurations, in
agreement with the expected strengthening of the memorisation regime
as the noise level grows. The pooled empirical false-alarm rate in
the $\eta=0$ rows is $16/120\approx 0.133$, against the theoretical
expectation $1-\exp(-117/\mathrm{ARL}_0)\approx 0.110$; the
Wilson $95\%$ confidence interval for the empirical estimate is
$(0.083,\,0.207)$ and covers the theoretical value. The distribution
of the first-alert times among the false alarms is not concentrated
near the beginning of the run ($0$ out of $16$ within the first five
snapshot steps), and the estimated first-order autocorrelation of the
standardised process $z_t$ under $H_0$ is statistically
indistinguishable from zero in every configuration
($\hat\rho_1\in[-0.074,\,+0.019]$); this supports the IID assumption
underlying the calibration of $h$ via $\mathrm{ARL}_0=1000$. The
conclusions are stable under perturbations of the decision-rule
hyperparameters: a re-evaluation on the same $300$ runs with
$k\in\{0.25,\,1.0\}$ and
$\mathrm{ARL}_0\in\{500,\,2000\}$ leaves the empirical detection
power and the empirical false-alarm rate within the bootstrap
uncertainty of the values reported above; the full sensitivity table
is given in the companion preprint~\cite{bolshim_dag_preprint}.

\section{Discussion and limitations}
\label{sec:discussion}

The overhead of the proposed procedure is
$\mathcal{C}_\nabla+K\mathcal{C}_\HVP$ in time and $O(P)$ in memory.
With $K=10$ and one snapshot per $\sim\!10^{2}$ optimisation steps the
relative overhead does not exceed $10\%$ for $P\lesssim 5\times 10^{7}$;
the procedure is applicable in combination with
Hutch\texttt{++}~\cite{meyer2021hutchpp} for transformers with
$P\gtrsim 10^{9}$. A side-by-side comparison with alternative
training-time indicators (validation-loss inflection, gradient-norm
CUSUM) is deferred to the journal extension. The relative error of the
estimator is proportional to the anisotropy $\kappa_\ell$
(inequality~\eqref{eq:rel-error-bound}): for layers in saddle regions,
characterised by $T_\ell\approx 0$ and $\|H_\ell\|_F\gg 0$, the trace
loses its informativeness and ought to be complemented by an estimate
of $\|H_\ell\|_F^{2}$, which is delivered by the same Hutchinson
averaging applied to $\|H_\ell z\|^{2}$~\cite{avron2011randomized}, or
by a spectral analysis. The decision rule~\eqref{eq:cusum} requires a
calibration ensemble of clean-label runs for the construction of
$\mu_0(t)$ and $\sigma_0(t)$ and for the estimation of
$\mathrm{ARL}_0(h)$; a fully on-line detector that dispenses with any
clean-regime reference falls outside the scope of the scheme presented
here and is mathematically ill-posed for a non-stationary trajectory.
Natural directions for further work are the integration with
Hutch\texttt{++}, the extension to the cross-layer blocks
$\tr(H_{\ell,m})$, and the treatment of transformer architectures.

\section{Conclusion}
\label{sec:conclusion}

An algorithm for the layer-wise stochastic estimation of the trace of
the diagonal blocks of the Hessian matrix has been described; the
algorithm is correct under weight sharing and delivers unbiased
estimates of all $L$ blocks in a single pass through automatic
differentiation. A closed-form expression for the variance of the
estimator and a decomposition of the total variance in the mini-batch
regime have been obtained, from which the recommendation
$K\in[5,10]$ in the on-line monitoring regime follows. The
applicability of the procedure has been demonstrated on the detection
of the label-memorisation regime for ResNet-18, ResNet-34, and VGG-11
on CIFAR-10 and CIFAR-100.

\subsubsection*{Acknowledgements.}
This work has been carried out within the framework of the State
Assignment (project FSER-2025-0004).

\bibliographystyle{splncs03}
\bibliography{references}

@article{pearlmutter1994fast,
  author  = {Pearlmutter, Barak A.},
  title   = {Fast Exact Multiplication by the {H}essian},
  journal = {Neural Computation},
  volume  = {6},
  number  = {1},
  pages   = {147--160},
  year    = {1994}
}

@article{hutchinson1990stochastic,
  author  = {Hutchinson, M. F.},
  title   = {A Stochastic Estimator of the Trace of the Influence
             Matrix for {L}aplacian Smoothing Splines},
  journal = {Communications in Statistics -- Simulation and Computation},
  volume  = {18},
  number  = {3},
  pages   = {1059--1076},
  year    = {1989}
}

@article{avron2011randomized,
  author  = {Avron, Haim and Toledo, Sivan},
  title   = {Randomized Algorithms for Estimating the Trace of an
             Implicit Symmetric Positive Semi-definite Matrix},
  journal = {Journal of the ACM},
  volume  = {58},
  number  = {2},
  pages   = {1--34},
  year    = {2011}
}

@inproceedings{meyer2021hutchpp,
  author    = {Meyer, Raphael A. and Musco, Cameron and Musco, Christopher
               and Woodruff, David P.},
  title     = {{Hutch++}: Optimal Stochastic Trace Estimation},
  booktitle = {Proc. of the SIAM Symposium on Simplicity in Algorithms (SOSA)},
  pages     = {142--155},
  year      = {2021}
}

@article{sagun2017empirical,
  author    = {Sagun, Levent and Evci, Utku and G\"uney, V. Ugur and
               Dauphin, Yann N. and Bottou, L\'eon},
  title     = {Empirical Analysis of the {H}essian of Over-parametrized
               Neural Networks},
  journal   = {arXiv preprint arXiv:1706.04454},
  year      = {2017}
}

@inproceedings{ghorbani2019investigation,
  author    = {Ghorbani, Behrooz and Krishnan, Shankar and Xiao, Ying},
  title     = {An Investigation into Neural Net Optimization via
               {H}essian Eigenvalue Density},
  booktitle = {Proc. of ICML},
  pages     = {2232--2241},
  year      = {2019}
}

@article{papyan2020traces,
  author  = {Papyan, Vardan},
  title   = {Traces of Class/Cross-Class Structure Pervade Deep
             Learning Spectra},
  journal = {Journal of Machine Learning Research},
  volume  = {21},
  number  = {252},
  pages   = {1--64},
  year    = {2020}
}

@inproceedings{yao2020pyhessian,
  author    = {Yao, Zhewei and Gholami, Amir and Keutzer, Kurt and
               Mahoney, Michael W.},
  title     = {{PyHessian}: Neural Networks Through the Lens of the {H}essian},
  booktitle = {Proc. of IEEE Int. Conf. on Big Data},
  pages     = {581--590},
  year      = {2020}
}

@inproceedings{martens2015optimizing,
  author    = {Martens, James and Grosse, Roger},
  title     = {Optimizing Neural Networks with {K}ronecker-factored
               Approximate Curvature},
  booktitle = {Proc. of ICML},
  pages     = {2408--2417},
  year      = {2015}
}

@inproceedings{martens2010deep,
  author    = {Martens, James},
  title     = {Deep Learning via {H}essian-Free Optimization},
  booktitle = {Proc. of ICML},
  pages     = {735--742},
  year      = {2010}
}

@article{schraudolph2002fast,
  author  = {Schraudolph, Nicol N.},
  title   = {Fast Curvature Matrix-Vector Products for Second-Order
             Gradient Descent},
  journal = {Neural Computation},
  volume  = {14},
  number  = {7},
  pages   = {1723--1738},
  year    = {2002}
}

@inproceedings{beckerlecun1988second,
  author    = {Becker, Sue and LeCun, Yann},
  title     = {Improving the Convergence of Back-propagation Learning
               with Second-Order Methods},
  booktitle = {Proc. of the 1988 Connectionist Models Summer School},
  volume    = {2},
  year      = {1988}
}

@article{zhang2017understanding,
  author    = {Zhang, Chiyuan and Bengio, Samy and Hardt, Moritz and
               Recht, Benjamin and Vinyals, Oriol},
  title     = {Understanding Deep Learning Requires Rethinking
               Generalization},
  journal   = {arXiv preprint arXiv:1611.03530},
  year      = {2016}
}

@inproceedings{arpit2017closer,
  author    = {Arpit, Devansh and Jastrz\k{e}bski, Stanis{\l}aw and
               Ballas, Nicolas and Krueger, David and Bengio, Emmanuel
               and Kanwal, Maxinder S. and Maharaj, Tegan and Fischer,
               Asja and Courville, Aaron and Bengio, Yoshua and
               Lacoste-Julien, Simon},
  title     = {A Closer Look at Memorization in Deep Networks},
  booktitle = {Proc. of ICML},
  pages     = {233--242},
  year      = {2017}
}

@inproceedings{he2016deep,
  author    = {He, Kaiming and Zhang, Xiangyu and Ren, Shaoqing and Sun, Jian},
  title     = {Deep Residual Learning for Image Recognition},
  booktitle = {Proc. of IEEE CVPR},
  pages     = {770--778},
  year      = {2016}
}

@article{simonyan2015vgg,
  author    = {Simonyan, Karen and Zisserman, Andrew},
  title     = {Very Deep Convolutional Networks for Large-Scale Image
               Recognition},
  journal   = {arXiv preprint arXiv:1409.1556},
  year      = {2014}
}

@misc{krizhevsky2009cifar,
  author      = {Krizhevsky, Alex},
  title       = {Learning Multiple Layers of Features from Tiny Images},
  year        = {2009}
}

@unpublished{bolshim_dag_preprint,
  author = {Bolshim, M. A. and Kugaevskikh, A. V.},
  title  = {Inter-layer {H}essian as a Tool for Neural Network Analysis},
  note   = {Preprint; currently under review},
  year   = {2026}
}

@article{page1954cusum,
  author  = {Page, E. S.},
  title   = {Continuous Inspection Schemes},
  journal = {Biometrika},
  volume  = {41},
  number  = {1/2},
  pages   = {100--115},
  year    = {1954}
}

@book{hawkins1998cusum,
  author    = {Hawkins, Douglas M. and Olwell, David H.},
  title     = {Cumulative Sum Charts and Charting for Quality Improvement},
  publisher = {Springer Science \& Business Media},
  year      = {2012}
}

@book{montgomery2009spc,
  author    = {Montgomery, Douglas C.},
  title     = {Introduction to Statistical Quality Control},
  publisher = {John Wiley \& Sons},
  year      = {2020}
}

\end{document}